\documentclass[11pt,letterpaper]{article}
\usepackage{naaclhlt2015}
\usepackage{amsmath}
\usepackage{amssymb}
\usepackage{times}
\usepackage{booktabs}
\usepackage{graphicx}
\usepackage{color}
\usepackage{latexsym}
\usepackage{caption}
\usepackage{subcaption}
\usepackage{sidecap}
\usepackage{soul}

\newcommand{\ignore}[1]{}

\newcommand{\obs}{w}
\newcommand{\obss}{\boldsymbol{\obs}}
\newcommand{\obsdomain}{V}
\newcommand{\rec}{\hat{w}}
\newcommand{\recs}{\boldsymbol{\rec}}

\newcommand{\hid}{t}
\newcommand{\hids}{\boldsymbol{\hid}}
\newcommand{\hiddomain}{T}
\newcommand{\emb}{\mathbf{v}}

\newenvironment{itemizesquish}{\begin{list}{\labelitemi}{\setlength{\itemsep}{0em}\setlength{\labelwidth}{0.5em}\setlength{\leftmargin}{\labelwidth}\addtolength{\leftmargin}{\labelsep}}}{\end{list}}

\title{Unsupervised POS Induction with Word Embeddings}

\author{Chu-Cheng Lin \qquad Waleed Ammar \qquad Chris Dyer \qquad Lori Levin\\
  	School of Computer Science\\
  	Carnegie Mellon University\\
  {\tt \{chuchenl,wammar,cdyer,lsl\}@cs.cmu.edu}
}

\date{}

\begin{document}
\maketitle
\begin{abstract}
Unsupervised word embeddings have been shown to be valuable as features in supervised learning problems; however, their role in unsupervised problems has been less thoroughly explored. In this paper, we show that embeddings can likewise add value to the problem of unsupervised POS induction. In two representative models of POS induction, we replace multinomial distributions over the vocabulary with multivariate Gaussian distributions over word embeddings and observe consistent improvements in eight languages. We also analyze the effect of various choices while inducing word embeddings on ``downstream'' POS induction results.
\end{abstract}

\section{Introduction}
Unsupervised POS induction is the problem of assigning word tokens to syntactic categories given only a corpus of untagged text. In this paper we explore the effect of replacing words with their vector space embeddings\footnote{Unlike \newcite{yatbaz:2014}, we leverage easily obtainable and widely used embeddings of word types.} in two POS induction models: the classic first-order HMM \cite{kupiec:92} and the newly introduced conditional random field autoencoder \cite{ammar:14}. 
In each model, instead of using a conditional \ul{multinomial distribution}\footnote{Also known as a categorical distribution.}\ul{ to generate a word token} $w_i \in V$ given a POS tag $t_i \in T$, we use a conditional \ul{Gaussian distribution and generate a $d$-dimensional word embedding} $\textbf{v}_{w_i} \in \mathbb{R}^d$ given $t_i$.

Our findings suggest that, in both models, substantial improvements are possible when word embeddings are used rather than opaque word types. 
However, the independence assumptions made by the model used to induce embeddings strongly determines its effectiveness for POS induction: embedding models that model \ul{short-range context are more effective} than those that model longer-range contexts.
This result is unsurprising, but it illustrates the lack of an evaluation metric that measures the syntactic (rather than semantic) information in word embeddings.
Our results also confirm the conclusions of \newcite{sirtis:2014} who were likewise able to improve POS induction results, albeit using a custom clustering model based on the the distance-dependent Chinese restaurant process \cite{blei:2011}.

Our contributions are as follows: 
(i) reparameterization of token-level POS induction models to use word embeddings; and
(ii) a systematic evaluation of word embeddings with respect to the syntactic information they contain.

\section{Vector Space Word Embeddings}

Word embeddings represent words in a language's vocabulary as points in a $d$-dimensional space such that nearby words (points) are similar in terms of their distributional properties. 
A variety of techniques for learning embeddings have been proposed, e.g., matrix factorization \cite{deerwester:90,dhillon:11} and neural language modeling \cite{mikolov:11,collobert:08}.

For the POS induction task, we specifically need embeddings that capture syntactic similarities. 
Therefore we experiment with two types of embeddings that are known for such properties:
\begin{itemizesquish}
\item \textbf{Skip-gram embeddings} \cite{mikolov-2:13} are based on a log bilinear model that predicts an unordered set of context words given a target word. 
\newcite{bansal:14} found that smaller context window sizes tend to result in embeddings with more syntactic information. We confirm this finding in our experiments.
\item \textbf{Structured skip-gram embeddings} \cite{ling:2015:naacl} extend the \textit{standard} skip-gram embeddings \cite{mikolov-2:13} by taking into account the relative positions of words in a given context.
\end{itemizesquish}
We use the tool \texttt{word2vec}\footnote{\texttt{https://code.google.com/p/word2vec/}} and \newcite{ling:2015:naacl}'s modified version\footnote{\texttt{https://github.com/wlin12/wang2vec/}} to generate both plain and structured skip-gram embeddings in nine languages.

\section{Models for POS Induction}
In this section, we briefly review two classes of models used for POS induction (HMMs and CRF autoencoders), and explain how to generate word embedding observations in each class.
We will represent a sentence of length $\ell$ as $\obss = \langle w_1, w_2, \ldots, \obs_{\ell} \rangle \in \obsdomain^{\ell}$ and a sequence of tags as $\hids = \langle \hid_1, \hid_2, \ldots, \hid_{\ell} \rangle \in \hiddomain^{\ell}$. The embeddings of word type $\obs \in \obsdomain$ will be written as $\emb_{\obs} \in \mathbb{R}^d$.
\subsection{Hidden Markov Models}
\label{sec:multinomial}
The hidden Markov model with multinomial emissions is a classic model for POS induction. This model makes the assumption that a latent Markov process with discrete states representing POS categories emits individual words in the vocabulary according to state (i.e., tag) specific emission distributions. An HMM thus defines the following joint distribution over sequences of observations and tags:
\begin{align}
p(\obss, \hids) = \prod_{i=1}^{\ell} p(\hid_i \mid \hid_{i-1}) \times p(\obs_i \mid \hid_i)
\label{eq:hmm}
\end{align}
where distributions $p(\hid_i \mid \hid_{i-1})$ represents the transition probability and $p(\obs_i \mid \hid_i)$ is the emission probability, the probability of a particular tag generating the word at position $i$.\footnote{Terms for the starting and stopping transition probabilities are omitted for brevity.}

We consider two variants of the HMM as baselines: 
\begin{itemizesquish}
\item $p(\obs_i \mid \hid_i)$ is parameterized as a ``na\"{\i}ve multinomial'' distribution with one distinct parameter for each word type.
\item $p(\obs_i \mid \hid_i)$ is parameterized as a multinomial logistic regression model with hand-engineered features as detailed in \cite{berg-kirkpatrick:10}.
\end{itemizesquish}
\paragraph{Gaussian Emissions.} 
\label{sec:gaussian}
We now consider  incorporating word embeddings in the HMM. 
Given a tag $t \in T$, instead of generating the observed word $w \in V$, we generate the (pre-trained) embedding $\mathbf{v}_{w} \in \mathbb{R}^d$ of that word. 
The conditional probability density assigned to $\mathbf{v}_w \mid t$ follows a multivariate Gaussian distribution with mean $\boldsymbol{\mu}_t$ and covariance matrix $\boldsymbol{\Sigma}_t$:
\begin{align}
p(\mathbf{v}_w; \boldsymbol{\mu}_t, \boldsymbol{\Sigma}_t) = \frac{
\exp\left(-\frac{1}{2}({\mathbf v}_w-{\boldsymbol\mu}_t)^{\top}{\boldsymbol\Sigma}_t^{-1}({\mathbf{v}_w}-{\boldsymbol\mu}_t)
\right)}{\sqrt{(2\pi)^d|\boldsymbol\Sigma_t|}}
\label{eq:gaussian}
\end{align}
This parameterization makes the assumption that embeddings of words which are often tagged as $t$ are concentrated around some point $\boldsymbol{\mu}_t \in \mathbb{R}^d$, and the concentration decays according to the covariance matrix $\boldsymbol{\Sigma}_t$.\footnote{``essentially, all models are wrong, but some are useful'' -- George E. P. Box}

Now, the joint distribution over a sequence of observations $\emb = \langle \emb_{\obs_1}, \emb_{\obs_2} \ldots, \emb_{\obs_\ell} \rangle$ (which corresponds to word sequence  $\obss = \langle {\obs_1}, {\obs_2}, \ldots, {\obs_\ell}, \rangle$) and a tag sequence $\boldsymbol{t} = \langle \hid_{1}, \hid_{2} \ldots, \hid_{\ell} \rangle$ becomes:
\begin{align*}
p(\emb, \hids) &= \prod_{i=1}^{\ell} p(\hid_i \mid \hid_{i-1}) \times p(\emb_{\obs_i} ; \boldsymbol{\mu}_{\hid_i}, \boldsymbol{\Sigma}_{\hid_i})
\end{align*}

We use the Baum--Welch algorithm to fit the $\boldsymbol{\mu}_{\hid}$ and $\boldsymbol{\Sigma}_{\hid_i}$ parameters. 
In every iteration, we update $\boldsymbol{\mu}_{t^*}$ as follows:
\begin{align}
\boldsymbol{\mu}_{t^*}^{\textit{new}} = \frac{\sum_{\emb\in {\cal{T}}} \sum_{i=1\ldots\ell} p(\hid_i=\hid^*\mid \emb)\times \emb_{\obs_i}}{\sum_{\emb\in {\cal{T}}} \sum_{i=1\ldots\ell} p(\hid_i=\hid^*\mid \emb) } \label{eq:baumwelch}
\end{align}
where $\cal{T}$ is a data set of word embedding sequences $\emb$ each of length $|\mathbf{v}|=\ell$, and $p(t_i=t^*\mid\emb)$ is the posterior probability of label $t^*$ at position $i$ in the sequence $\emb$.
Likewise the update to $\boldsymbol{\Sigma}_{t^*}$ is:
\begin{align}
\boldsymbol{\Sigma}_{t^*}^{\textit{new}} = \frac{\sum_{\emb\in {\cal{T}}} \sum_{i=1\ldots\ell} p(\hid_i=\hid^*\mid \emb)\times \boldsymbol{\delta} \boldsymbol{\delta}^{\top}  }{\sum_{\emb\in {\cal{T}}} \sum_{i=1\ldots\ell} p(\hid_i=\hid^*\mid \emb) } \label{eq:baumwelch-covar}
\end{align}
where $\boldsymbol{\delta}=\emb_{\obs_i}-\boldsymbol{\mu}_{t^*}^{\textit{new}}$.

\subsection{Conditional Random Field Autoencoders}
\label{sec:gaussianautoencoder}
The second class of models this work extends is called CRF autoencoders, which we recently proposed in \cite{ammar:14}.
It is a scalable family of models for feature-rich learning from unlabeled examples. 
The model conditions on one copy of the structured input, and generates a reconstruction of the input via a set of interdependent latent variables which represent the linguistic structure of interest. 
As shown in Eq.~\ref{eq:crfautoencoder}, the model factorizes into two distinct parts: the encoding model $p(\hids\mid\obss)$ and the reconstruction model $p(\recs\mid\hids)$; where $\obss$ is the structured input (e.g., a token sequence), $\hids$ is the linguistic structure of interest (e.g., a sequence of POS tags), and $\recs$ is a generic reconstruction of the input. 
For POS induction, the encoding model is a linear-chain CRF with feature vector $\boldsymbol{\lambda}$ and local feature functions $\mathbf{f}$.
\begin{align} 
p(\recs,& \hids\mid \obss) = p(\hids\mid\obss) \times p(\recs\mid\hids)  \nonumber \\ 
&\propto p(\recs\mid\hids) \times \exp \boldsymbol{\lambda} \cdot \sum_{i=1}^{|\obss|} \mathbf{f}(\hid_i, \hid_{i-1}, \obss) \label{eq:crfautoencoder}
\end{align}

In \cite{ammar:14}, we explored two kinds of reconstructions $\recs$: surface forms and Brown clusters \cite{brown:92}, and used ``stupid multinomials'' as the underlying distributions for re-generating $\recs$.

\paragraph{Gaussian Reconstruction.}
In this paper, we use $d$-dimensional word embedding reconstructions $\rec_i=\textbf{v}_{w_i} \in \mathbb{R}^d$, and replace the multinomial distribution of the reconstruction model with the multivariate Gaussian distribution in Eq.~\ref{eq:gaussian}. 
We again use the Baum--Welch algorithm to estimate $\boldsymbol{\mu}_{t^*}$ and $\boldsymbol{\Sigma}_{t^*}$ similar to Eq.~\ref{eq:baumwelch}. 
The only difference is that posterior label probabilities are now conditional on both the input sequence $\obss$ and the embeddings sequence $\mathbf{v}$, i.e., replace $p(t_i=t^*\mid \mathbf{v})$ in Eq.~\ref{eq:gaussian} with $p(t_i=t^*\mid \obss, \mathbf{v})$.

\section{Experiments}
\label{sec:experiments}

\begin{figure*}[]
  \includegraphics[width=.53\linewidth]{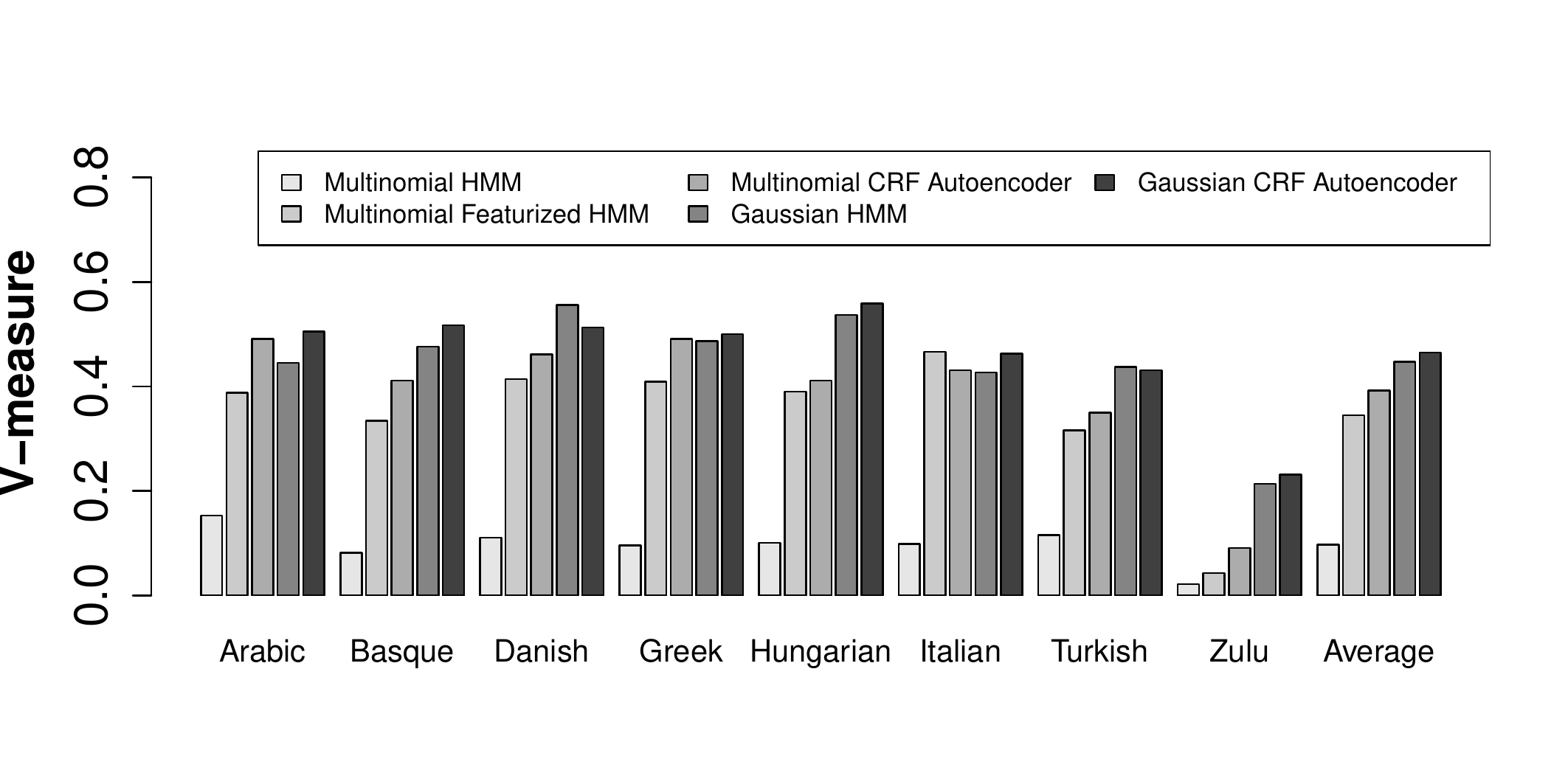}
  \includegraphics[width=.53\linewidth]{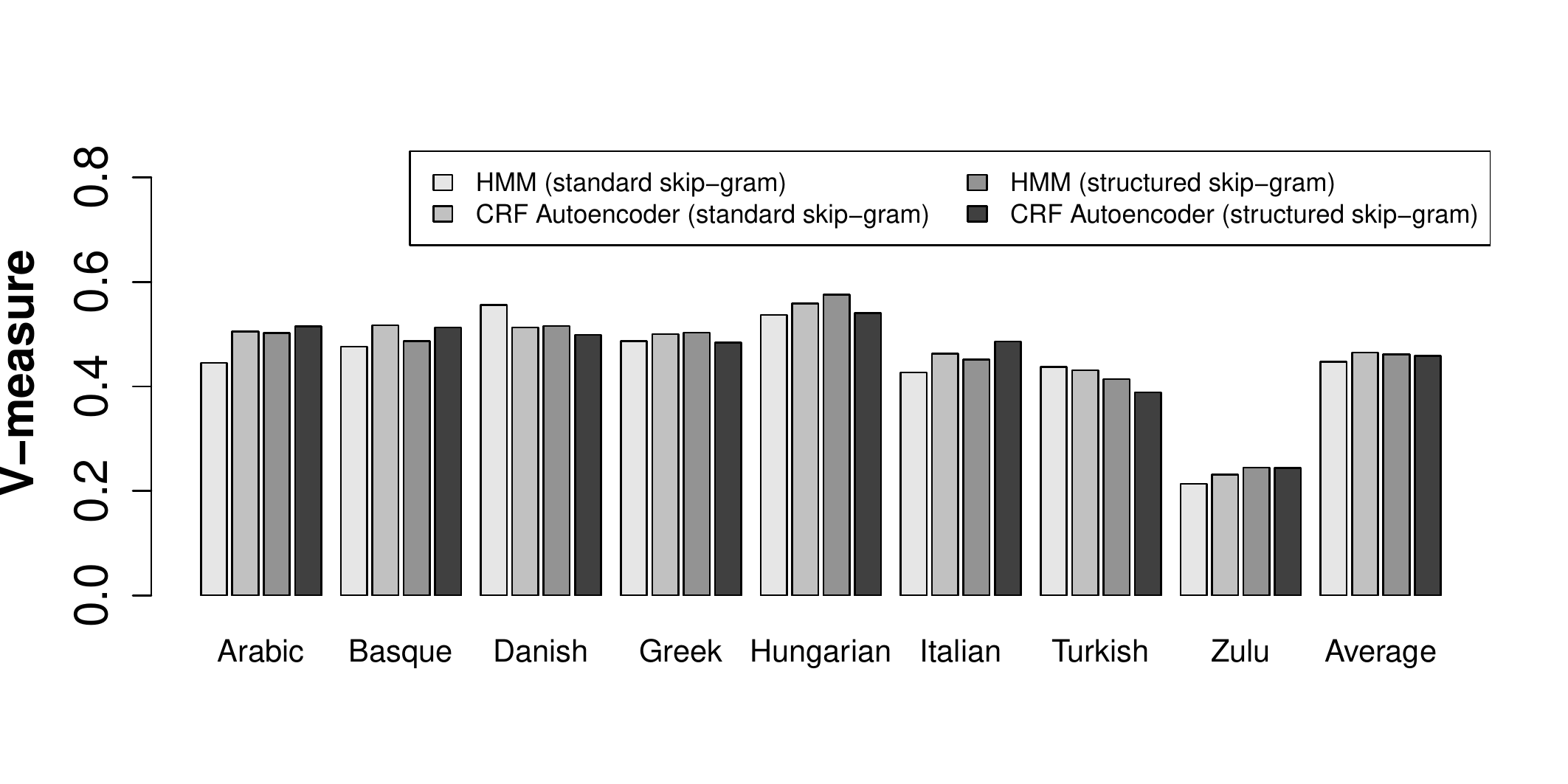}
  \caption{
POS induction results. (V-measure, higher is better.) Window size is $1$ for all word embeddings. \textbf{Left}: Models which use standard skip-gram word embeddings (i.e., Gaussian HMM and Gaussian CRF Autoencoder) outperform all baselines on average across languages. 
\textbf{Right}: comparison between standard and structured skip-grams on Gaussian HMM and CRF Autoencoder.} \label{fig:vmeasure}\label{fig:vmeasure_sg_ssg}
\end{figure*}

In this section, we attempt to answer the following questions:
\begin{itemizesquish}
\item \S\ref{sec:choiceofmodels}: Do syntactically-informed word embeddings improve POS induction? Which model performs best?
\item \S\ref{sec:choiceofembeddings}: What kind of word embeddings are suitable for POS induction?
\end{itemizesquish}

\subsection{Choice of POS Induction Models}
\label{sec:choiceofmodels}
Here, we compare the following models for POS induction:
\begin{itemizesquish}
\item Baseline: HMM with multinomial emissions \cite{kupiec:92},
\item Baseline: HMM with log-linear emissions \cite{berg-kirkpatrick:10}, 
\item Baseline: CRF autoencoder with multinomial reconstructions \cite{ammar:14},\footnote{We use the configuration with best performance which reconstructs Brown clusters.}
\item Proposed: HMM with Gaussian emissions, and
\item Proposed: CRF autoencoder with Gaussian reconstructions.
\end{itemizesquish}

\paragraph{Data.}
To train the POS induction models, we used the plain text from the training sections of the CoNLL-X shared task \cite{buchholz:06} (for Danish and Turkish), the CoNLL 2007 shared task \cite{nivre:07} (for Arabic, Basque, Greek, Hungarian and Italian), and the Ukwabelana corpus \cite{spiegler:10} (for Zulu).
For evaluation, we obtain the corresponding gold-standard POS tags by deterministically mapping the language-specific POS tags in the aforementioned corpora to the corresponding universal POS tag set \cite{petrov:12}. 
This is the same set up we used in \cite{ammar:14}.

\paragraph{Setup.}
In this section, we used skip-gram (i.e., \texttt{word2vec}) embeddings with a context window size $=1$ and with dimensionality $d=100$, trained with the largest corpora for each language in \cite{quasthoff:06}, in addition to the plain text used to train the POS induction models.\footnote{We used the \texttt{corpus/tokenize-anything.sh} script in the cdec decoder \cite{dyer:10} to tokenize the corpora from \cite{quasthoff:06}. The other corpora were already tokenized. In Arabic and Italian, we found a lot of discrepancies between the tokenization used for inducing word embeddings and the tokenization used for evaluation. We expect our results to improve with consistent tokenization. }
In the proposed models, we only show results for estimating $\boldsymbol{\mu}_t$, assuming a diagonal covariance matrix $\boldsymbol{\Sigma}_t(k,k)=0.45 \forall k \in \{1,\ldots, d\}$.\footnote{Surprisingly, we found that estimating $\boldsymbol{\Sigma}_t$ significantly degrades the performance. This may be due to overfitting \cite{shinozaki:07}. Possible remedies include using a prior \cite{gauvain:94}.}
While the CRF autoencoder with multinomial reconstructions were carefully initialized as discussed in \cite{ammar:14}, CRF autoencoder with Gaussian reconstructions were initialized uniformly at random in $[-1,1]$. 
All HMM models were also randomly initialized.
We tuned all hyperparameters on the English PTB corpus, then fixed them for all languages.

\paragraph{Evaluation.}
We use the V-measure evaluation metric \cite{rosenberg:07} to evaluate the predicted syntactic classes at the token level.\footnote{We found the V-measure results to be consistent with the many-to-one evaluation metric \cite{johnson:07}. We only show one set of results for brevity.}

\paragraph{Results.}
The results in Fig.~\ref{fig:vmeasure} (left) clearly suggest that we can use word embeddings to improve POS induction. 
Surprisingly, the feature-less Gaussian HMM model outperforms the strong feature-rich baselines: Multinomial Featurized HMM and Multinomial CRF Autoencoder.
One explanation is that our word embeddings were induced using larger unlabeled corpora than those used to train the POS induction models.
The best results are obtained using both word embeddings and feature-rich models using the Gaussian CRF autoencoder model.
This set of results suggest that word embeddings and hand-engineered features play complementary roles in POS induction.
It is worth noting that the CRF autoencoder model with Gaussian reconstructions did not require careful initialization.\footnote{In \cite{ammar:14}, we found that careful initialization for the CRF autoencoder model with multinomial reconstructions is necessary.}

\subsection{Choice of Embeddings}
\label{sec:choiceofembeddings}

\paragraph{Standard skip-gram vs. structured skip-gram.}
On Gaussian HMMs, structured skip-gram embeddings score moderately higher than standard skip-grams. And as the context window size gets larger the gap widens. The reason may be that structured skip-gram embeddings give each position within the context window its own project matrix, so the smearing effect is not as pronounced as the window grows when compared to the standard embeddings. However the best performance is still obtained when the window is small.\footnote{In preliminary experiments, we also compared standard skip-gram embeddings to SENNA embeddings \cite{collobert:11} (which are trained in a semi-supervised multi-task learning setup, with one task being POS tagging) on a subset of the English PTB corpus. As expected, the induced POS tags are much better when using SENNA embeddings, yielding a V-measure score of $0.57$ compared to $0.51$ for skip-gram embeddings. Since SENNA embeddings are only available in English, we did not include it in the comparison in Fig.~\ref{fig:vmeasure}. }

\begin{figure}[t]
\centering
\includegraphics[width=\linewidth]{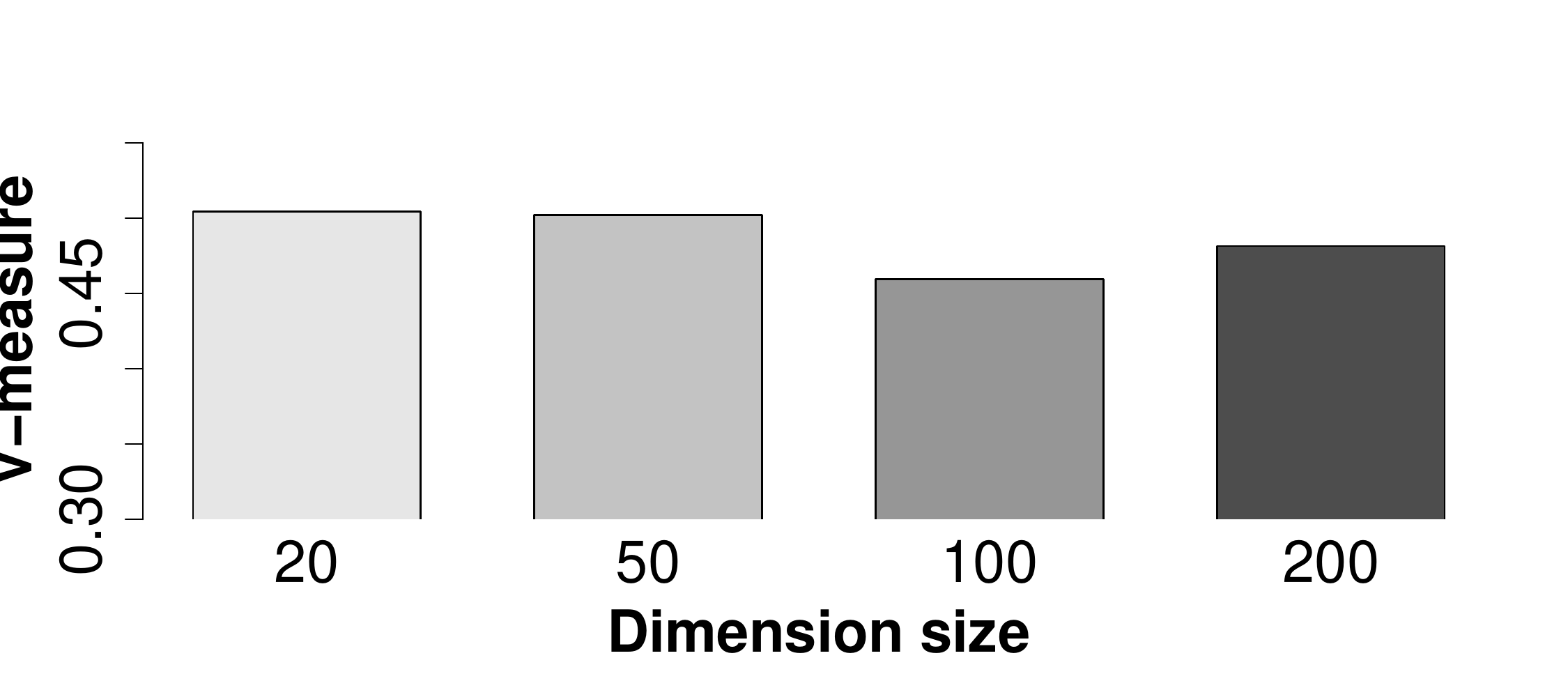}
\caption{Effect of dimension size on POS induction on a subset of the English PTB corpus. $w = 1$. The model is HMM with Gaussian emissions.\label{fig:hmm-dim-size}}
\end{figure}

\paragraph{Dimensions $\mathbf{=20}$ vs. $\mathbf{200}$.}
We also varied the number of dimensions in the word vectors $(d\in\{20, 50, 100, 200\})$. The best V-measure we obtain is $0.504$ ($d=20$) and the worst is $0.460$ ($d=100$).
However, we did not observe a consistent pattern as shown in Fig.~\ref{fig:hmm-dim-size}.

\begin{figure}[t]
\centering
\includegraphics[width=\linewidth]{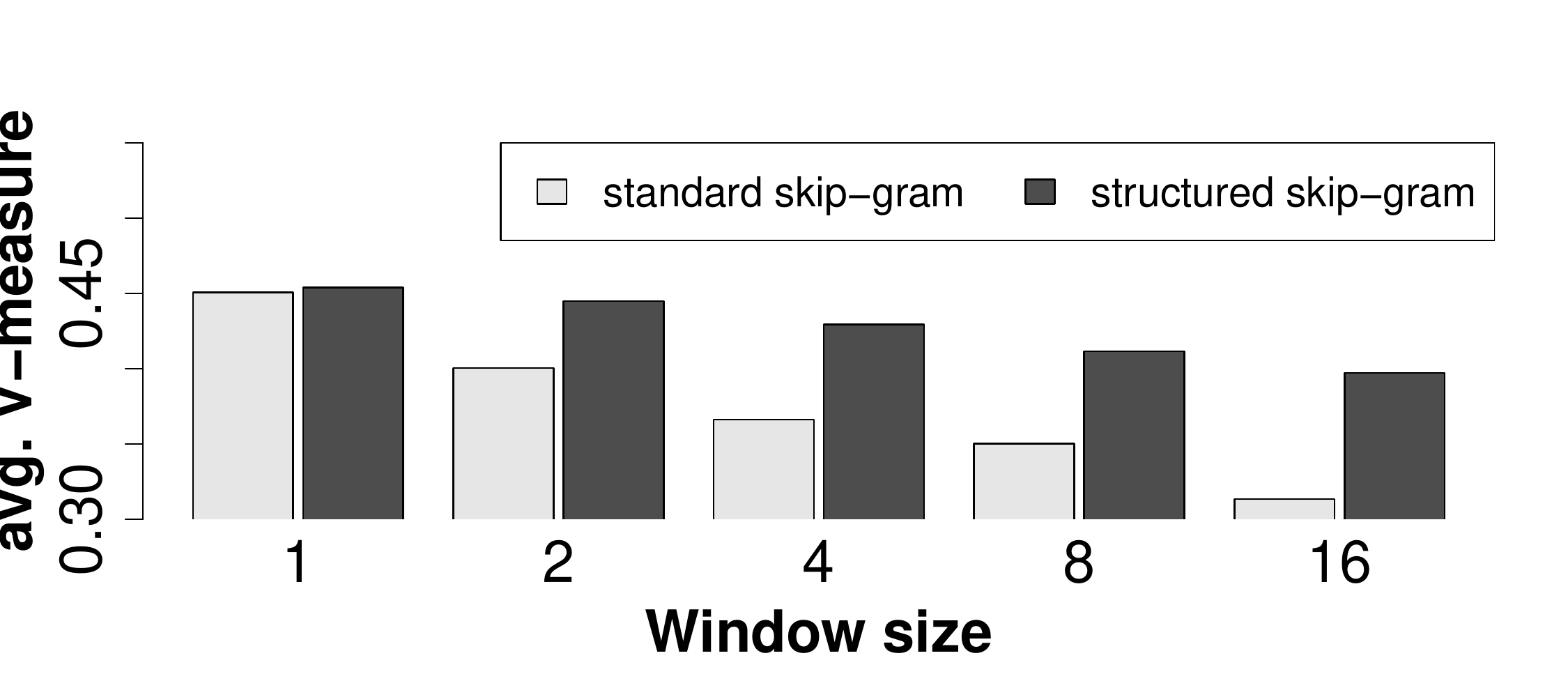}
\caption{
Effect of window size and embeddings type on POS induction over the languages in Fig.~\ref{fig:vmeasure}. $d=100$. The model is HMM with Gaussian emissions. \label{fig:hmm-wsize-avg-vm}}
\end{figure}

\paragraph{Window size $\mathbf{=1}$ vs. $\mathbf{16}$.}
Finally, we varied the window size for the context surrounding target words $(w\in\{1,2,4,8,16\})$.
$w=1$ yields the best average V-measure across the eight languages as shown in Fig.~\ref{fig:hmm-wsize-avg-vm}. 
This is true for both standard and structured skip-gram models.
Notably, larger window sizes appear to produce word embeddings with less syntactic information.
This result confirms the observations of \newcite{bansal:14}.

\subsection{Discussion}
We have shown that (re)generating word embeddings does much better than generating opaque word types in unsupervised POS induction. At a high level, this confirms prior findings that unsupervised word embeddings  capture syntactic properties of words, and shows that different embeddings capture more syntactically salient information than others. As such, we contend that unsupervised POS induction can be seen as a diagnostic metric for assessing the syntactic quality of embeddings.

To get a better understanding of what the multivariate Gaussian models have learned, we conduct a hill-climbing experiment on our English dataset. We seed each POS category with the average vector of 10 randomly sampled words from that category and train the model. Seeding unsurprisingly improves tagging performance. We also find words that are the nearest to the centroids generally agree with the correct category label, which validate our assumption that syntactically similar words tend to cluster in the high-dimensional embedding space. It also shows that careful initialization of model parameters can bring further improvements.

However we also find that words that are close to the centroid are not
necessarily representative of what linguists consider to be prototypical.
For example, \newcite{hopper:83} show that physical, telic, past tense
verbs are more prototypical with respect to case marking, agreement, and
other syntactic behavior.   However, the verbs nearest our centroid  all
seem rather abstract. In English, the nearest 5 words in the verb category are \emph{entails, aspires, attaches, foresees, deems}.
 This
may be because these words seldom serve functions other than verbs; and
placing the centroid around them incurs less penalty (in contrast to physical verbs, e.g. \emph{bite}, which often also act as nouns).
Therefore one should be cautious in interpreting what is prototypical
about them.

\ignore{
\begin{table}
\begin{tabular}{c}
\toprule
entails \\
aspires \\
attaches \\
foresees \\
deems \\
\bottomrule
\end{tabular}
\caption{5 words nearest to the centroid of verbs. Structured skip-gram embeddings are used. Ill-formed words due to improper tokenization are filtered.}
\label{tbl:verb-centroid}
\end{table}
}

\ignore{of the reasons is the continuous representations allows sharing of statistical strength. Say after an E--step during the Gaussian HMM inference, we have $\mathbb{E}_{\hid}(\textrm{apple})=90$, $\mathbb{E}_{\hid}(\textrm{orange})=5$, and $\mathbb{E}_{\hid}(\textrm{likely})=5$. Then the M--step following it would assign
$p(\textrm{apple} \mid \hid)=0.9$, $p(\textrm{orange} \mid \hid)=p(\textrm{likely} \mid \hid)=0.05$ despite (assumed) similarity between `orange' and `apple'. On the other hand, in the word embedding regenerating models, the MLE of $\mu_{\hid}$ would be a weighted average of $\emb_{\obs}$; and $p( \textrm{orange} \mid \mu_{\hid}) > p( \textrm{likely} \mid \mu_{\hid})$ if $\|\emb_{\textrm{likely}}-\emb_{\textrm{apple}}\| > \|\emb_{\textrm{orange}}-\emb_{\textrm{apple}}\|$.

It is possible to explicitly model the context similarity via additional latent variables\cite{matsuzaki:05}, but such models would be computationally much expensive than a simple Gaussian HMM that relies upon learned embeddings.
}

\section{Conclusion}
We propose using a multivariate Gaussian model to generate vector space representations of observed words in generative or hybrid models for POS induction, as a superior alternative to using multinomial distributions to generate categorical word types. 
We find the performance from a simple Gaussian HMM competitive with strong feature-rich baselines. We further show that substituting the emission part of the CRF autoencoder can bring further improvements.
We also confirm previous findings which suggest that smaller context windows in skip-gram models result in word embeddings which encode more syntactic information.
It would be interesting to see if we can apply this approach to other tasks which require generative modeling of textual observations such as language modeling and grammar induction.

\ignore{For future work, we found that the choice of the covariance $\Sigma_{\hid}$ is crucial, suggesting that better generative models of the embedding space can further boost the performance.}

\section*{Acknowledgements}{
This work was sponsored by the U.S.~Army Research Laboratory and the U.S.~Army Research Office under contract/grant numbers  W911NF-11-2-0042 and W911NF-10-1-0533.
The statements made herein are solely the responsibility of the authors.
}

\bibliographystyle{naaclhlt2015}
\bibliography{biblio}

\end{document}